%% file: main_v2_arxiv.tex
\documentclass{article}
\usepackage{iclr2027_conference,times}
\iclrfinalcopy

\input{math_commands.tex}

\usepackage{graphicx}
\usepackage{caption}
\usepackage{subcaption}
\usepackage{booktabs}
\usepackage{multirow}
\usepackage{array}
\usepackage{makecell}
\usepackage{siunitx}
\usepackage{microtype}
\usepackage{hyperref}
\usepackage{url}
\usepackage{etoolbox}
\usepackage{wrapfig}
\usepackage{placeins}
\usepackage{float}
\usepackage{amsmath}
\usepackage{amssymb}
\usepackage{amsfonts}
\usepackage{bbm}
\usepackage{xspace}
\usepackage{xcolor}
\usepackage[most]{tcolorbox}

\newcommand{\floatbodycaptionskip}{\vspace{-4pt}}
\pretocmd{\caption}{\floatbodycaptionskip}{}{}

\newcommand{\method}{\textsc{ChartMind}\xspace}
\newcommand{\probe}{\textsc{ProbeBench}\xspace}
\newcommand{\stitle}[1]{\noindent\textbf{#1}}

\definecolor{tagblue}{RGB}{50,100,220}
\definecolor{tagred}{RGB}{220,50,50}
\definecolor{taggreen}{RGB}{50,140,60}
\newtcolorbox{sampleblock}[1]{
  enhanced,
  breakable,
  colback=black!2,
  colframe=black!28,
  boxrule=0.5pt,
  arc=1pt,
  left=7pt,
  right=7pt,
  top=5pt,
  bottom=5pt,
  before skip=6pt,
  after skip=7pt,
  title={#1},
  fonttitle=\bfseries\small,
  fontupper=\ttfamily\small\raggedright,
}
\newtcblisting{promptblock}[1]{
  enhanced,
  listing only,
  listing engine=listings,
  colback=black!2,
  colframe=black!28,
  boxrule=0.5pt,
  arc=1pt,
  left=7pt,
  right=7pt,
  top=5pt,
  bottom=5pt,
  before skip=6pt,
  after skip=7pt,
  title={#1},
  fonttitle=\bfseries\small,
  listing options={
    basicstyle=\ttfamily\footnotesize,
    breaklines=true,
    breakatwhitespace=true,
    columns=fullflexible,
    keepspaces=true,
    showstringspaces=false
  },
}
\newtcolorbox{caseblock}[1]{
  enhanced,
  colback=black!2,
  colframe=black!28,
  boxrule=0.5pt,
  arc=1pt,
  left=7pt,
  right=7pt,
  top=3pt,
  bottom=3pt,
  before skip=4pt,
  after skip=4pt,
  title={#1},
  fonttitle=\bfseries\small,
  fontupper=\ttfamily\footnotesize\raggedright,
}

\title{Charts as Epistemic Actions for Experience-Driven Agentic Data Analysis}

\author{
Xinzhe Wang\textsuperscript{1} \quad
Xudong Weng\textsuperscript{1} \quad
Zang Li\textsuperscript{1} \quad
Shaolei Zhang\textsuperscript{2}\\
\textsuperscript{1}Tencent \quad
\textsuperscript{2}Renmin University of China\\
\texttt{\{xinzhewang,steveweng,gavinzli\}@tencent.com}\\
\texttt{zhangshaolei98@ruc.edu.cn}
}

\begin{document}

\maketitle

\begin{abstract}
Data agents built on ReAct-style loops now execute complex analytical workflows over structured data, yet their intermediate reasoning still runs almost entirely through text. Human analysts instead move between numbers, tables, and charts, reading visual evidence to expose trends and anomalies that text summaries drop. We ask whether charts can serve as internal epistemic actions inside a data agent's reasoning loop, and how to teach an agent to use them. We first build \probe{}, a suite of 200 controlled tasks under four matched interface conditions, and find a persistent gap between what agents can read and what they choose to use. Agents score as well from self-generated charts as from numeric output, yet plotting access alone does not translate into consistent use. Task slices show that charts lead on temporal, pattern, and anomaly reasoning, while text leads on precise reading and conditional verification. We then propose \method{}, which distils two families of experience from conditioned trajectory rollouts. One family records how to select and verify analytical operations, and the other records when to invoke text, charts, or both. A trajectory judge scores outcomes and labels intermediate steps, and a validation gate admits only experience that improves held-out performance. On three held-out benchmarks with four LLM backbones, \method{} raises analytical accuracy and visual evidence quality over matched baselines, and backbones that never drew charts begin using them, all without touching model parameters. 
\end{abstract}
\vspace{-0.75ex}

\section{Introduction}
\label{sec:introduction}

\noindent Data agents translate an analytical goal into a sequence of grounded actions and derive insights step by step. Recent progress in ReAct-style reasoning and acting loops \citep{yao2023react} has enabled these agents to plan, execute, inspect, and revise increasingly complex workflows over structured data. Analytical report generation, long-horizon exploration, and exact-answer analysis all now sit within their reach \citep{lei2025dacomp,liu2026ddrbench,jing2024dsbench}. Such agents execute text-based analytical actions competently.

\noindent Human analysts, in contrast, do not think only in text. They move continually between numerical summaries, tables, and charts, using visual evidence to expose trends, distributions, and anomalies that a scrolling table hides. A well-chosen chart at the right moment can reshape the next step of analysis. Charts thus act as epistemic actions in the analyst's loop rather than as post-hoc report artefacts. We therefore ask whether an agent should also treat charts as intermediate evidence and learn when to invoke them.

\noindent Existing methods address only part of this decision. Chart-generation methods optimise visualisations requested by a user or an agent \citep{dibia2023lida,yang2024matplotagent,namgoong2025amace,lu2026multivisagent,wang2025chartoptimiser}, while ChartMimic benchmarks chart-to-code generation \citep{yang2025chartmimic}. Chart-understanding methods assume that a chart is already available and focus on extracting information from it \citep{masry2023unichart,han2023chartllama,liu2023deplot,liu2023matcha,zhang2024tinychart,kaur2026chartagent,das2025chartsofthought}, while ChartBench benchmarks complex chart reasoning \citep{xu2023chartbench}. Recent work compares different input representations \citep{zhou2025fres,ho2026formatmatters}, routes table--query pairs across textual, visual, and fused paths \citep{xing2026tabledart}, or builds visual evidence on demand during report writing \citep{lin2026evidfuse}. We study how a data agent selects textual and chart evidence at intermediate analytical states.

\noindent To turn this open question into a concrete study, we build \probe{}, a probe suite of 200 tasks synthesised from InsightBench \citep{sahu2025insightbench} with four matched interface conditions and a common ReAct backbone. Three findings emerge. First, charts alone already sustain end-to-end analysis. Second, plotting access alone does not translate into consistent use. Third, charts and text behave as complementary evidence across task slices, with charts stronger on temporal, pattern, and anomaly reasoning and text stronger on precise reading and conditional verification. These findings motivate a method that turns representation choice into a learned behaviour of the agent.

\noindent We propose \method{}, an experience-driven method that teaches a data agent to use charts inside the analytical loop. For each \probe{}, \method{} generates diverse trajectories under different representation conditions and uses $f_{\mathrm{score}}$ and $f_{\mathrm{attr}}$ to score outcomes and label intermediate decisions. A builder agent distils the labelled trajectories into two experience families. Analysis experience $\mathcal{E}_{\mathrm{ana}}$ records how to select and verify analytical operations, and representation experience $\mathcal{E}_{\mathrm{repr}}$ records when and how to invoke text, charts, or a mixture. Candidate packages are accepted only when a validation gate confirms that held-out performance strictly improves. On downstream tasks the data agent reads a frozen package as external memory, so representation choice becomes part of the analytical policy without changing model parameters. 

\noindent Our main contributions are as follows.
\begin{itemize}
\setlength{\itemsep}{2pt}
\setlength{\topsep}{4pt}
\setlength{\parsep}{0pt}
\item \textbf{Probe study of chart use.} We construct \probe{} with four matched interfaces and four backbones. The suite exposes a capability--use gap in current data agents and quantifies where charts and text serve as complementary evidence.
\item \textbf{Experience-driven method for chart use.} We propose \method{}. Given conditioned trajectory rollouts, it distils analysis and representation experience under outcome scoring and step-level attribution, and only experience updates that pass a validation gate enter the frozen package. \method{} is the first framework that treats charts as internal epistemic actions of a data agent through learned experience.
\item \textbf{Consistent gains on held-out benchmarks.} On DAComp-DA, DDR-Bench, and DSBench with four LLM backbones, \method{} raises analytical accuracy and visual evidence quality over matched baselines. Backbones with weak representation-selection habits begin using charts, and ablations confirm the contributions of rollout diversity, validation gating, and both experience types.
\end{itemize}

\section{Charts in the Analytical Loop}
\label{sec:probe}

\noindent This section asks whether charts carry genuine analytical value for data agents and how their utility differs from text. We design \probe{}, a controlled probe suite, and evaluate representative agents under four matched interfaces, then report the suite's construction, evaluation protocol, and three empirical findings.

\subsection{Probe Design}
\label{sec:probe-setup}

\stitle{Task construction.}\quad To isolate the effect of intermediate representations on agent reasoning, we build \probe{} from InsightBench \citep{sahu2025insightbench}, which provides enterprise tables, open-ended questions, executable analyses, and target insights. Each valid insight defines a task, yielding 200 tasks in total. A task contains a raw CSV, dataset description, analytical question, reference analysis, and target insight. The agent sees the table, question, description, schema, and first three rows, then selects and executes operations to answer in natural language; the target insight is hidden for evaluation.

\stitle{Analytical dimensions.}\quad For post-hoc slicing without label leakage, we annotate two dimensions. \textit{Question type} covers numeric lookup, composition, ranking, group comparison, temporal change, relationship, anomaly, diagnosis, and prediction. \textit{Cognitive load} covers information location, scan-and-select, comparison, pattern recognition, relational interpretation, multidimensional integration, and conditional reasoning. An LLM proposes the labels and a human verifies them; they never enter the agent context.

\stitle{Agent interfaces.}\quad To hold everything but the analytical interface constant, all four settings share a vanilla ReAct agent with one general-purpose Bash tool. Each call starts a fresh process, files persist in a shared working directory, and the CSV is exposed read-only. We fix the model, initial context, interaction limit, and tool budget across conditions. \textbf{No-tool} removes execution tools. \textbf{Text-only} exposes stdout and stderr but forbids plotting. \textbf{Chart-only} exposes generated images and stderr while hiding stdout. \textbf{Free-style} returns stdout, stderr, and images and allows both text and charts.

\stitle{Evaluation.}\quad To obtain comparable scores across settings, a shared LLM judge grades each answer against the target insight on a normalised $[0,100]$ scale. Slice analyses use task-paired differences with clustered bootstrap intervals \citep{brown1995bootstrap} and randomisation tests. Figure~\ref{fig:four-condition} and Table~\ref{tab:model-bias} summarise the two views used in the following analysis.

\begin{table}[t]
\centering
\begin{minipage}[t]{0.32\textwidth}
\centering
\includegraphics[width=\linewidth]{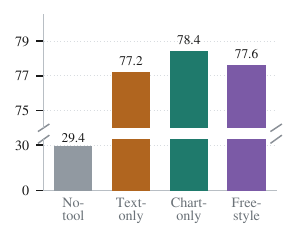}
\captionof{figure}{Judge scores under four matched \probe{} interfaces. The score axis is split.}
\label{fig:four-condition}
\end{minipage}\hfill
\begin{minipage}[t]{0.66\textwidth}
\centering
\small
\vspace{-31mm}
\captionof{table}{Representation choice under \textbf{Free-style} across four backbones. \textit{Both} counts tasks that use text and charts jointly.}
\label{tab:model-bias}
\setlength{\tabcolsep}{3.2pt}
\renewcommand{\arraystretch}{1.05}
\begin{tabular}{@{}lcccccc@{}}
\toprule
\multirow{2}{*}[-0.15em]{LLM backbone} &
\multirow{2}{*}[-0.15em]{Score} &
\multicolumn{3}{c}{Representation use} &
\multicolumn{2}{c}{Interaction cost} \\
\cmidrule(lr){3-5}\cmidrule(lr){6-7}
& & Text & Chart & Both & Rounds & Tokens \\
\midrule
Claude Sonnet 5     & 79.2 & 102 & 0 & 98  & 4.0 & 18.7k \\
GPT-5.6 Luna        & 76.2 & 198 & 0 & 2   & 3.1 & 6.8k  \\
GPT-5.6 Sol         & 77.6 & 198 & 0 & 2   & 2.9 & 6.4k  \\
DeepSeek V4.1 Flash & 80.2 & 143 & 0 & 57  & 6.2 & 30.1k \\
\bottomrule
\end{tabular}
\end{minipage}
\end{table}
\vspace{-4pt}

\subsection{Empirical Findings}
\label{sec:probe-findings}

\stitle{Finding 1. Charts alone already sustain end-to-end analysis.}\quad To check whether charts carry enough evidence for analytical questions, we compare the four interfaces under the same backbone. Figure~\ref{fig:four-condition} shows that removing execution tools collapses the agent to a low score, while the Text-only and Chart-only interfaces achieve similar performance. Chart-only slightly outperforms Text-only, so the agent can read charts it just drew and produce answers comparable in quality to those obtained from numeric tables. Charts already satisfy the sufficiency condition for intermediate evidence, so the remaining question is whether agents actually use them.

\stitle{Finding 2. Representation choice is model-dependent, not capability-driven.}\quad To probe whether the sufficiency of charts translates into use, we let each backbone freely choose between text and charts under the same Free-style interface. Table~\ref{tab:model-bias} shows large behavioural divergence. Claude Sonnet 5 combines text and charts on nearly half of the tasks and DeepSeek V4.1 Flash on more than a quarter, while GPT-5.6 Luna and GPT-5.6 Sol essentially fall back to text. The frequency of chart use does not track score. This shows that identical tool access yields inconsistent representation habits and confirms a capability--use gap. A method that closes this gap should shape when and how charts enter the loop, and cannot stop at exposing plotting tools.

\begin{figure}[t!]
	\centering
	\vspace{-2pt}
	\begin{subfigure}[t]{\linewidth}
		\centering
		\includegraphics[width=0.84\linewidth]{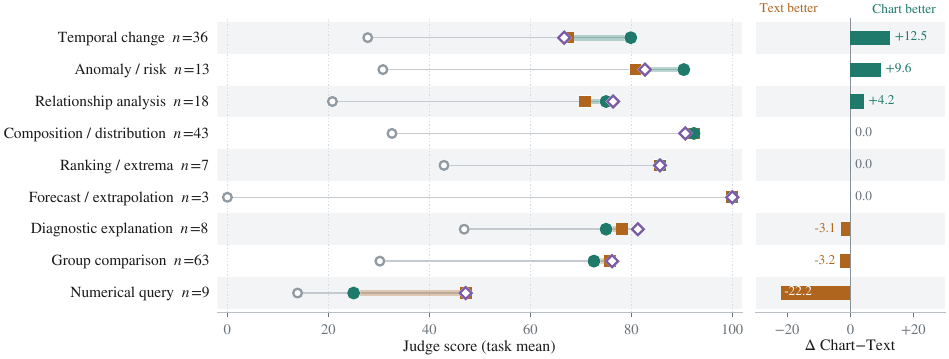}
	\end{subfigure}\\[-1pt]
	\begin{subfigure}[t]{\linewidth}
		\centering
		\includegraphics[width=0.84\linewidth]{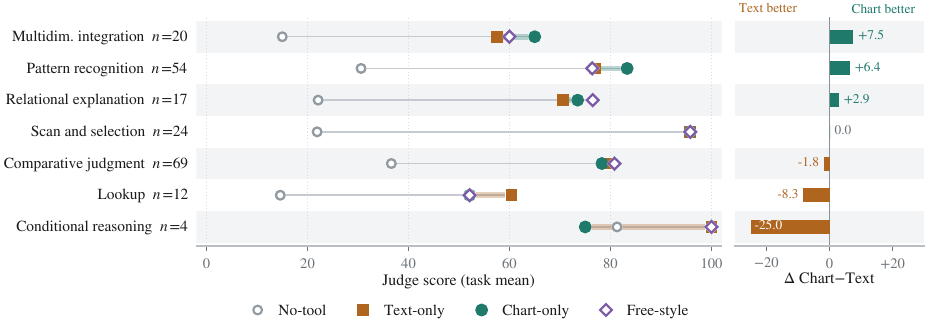}
	\end{subfigure}
	\caption{Slice analysis of the four \probe{} interfaces. \textbf{Top:} slices by question type. \textbf{Bottom:} slices by primary cognitive load. Markers show slice mean scores. The right column reports the Chart-only minus Text-only gap.}
	\label{fig:probe-slices}
	\vspace{-4pt}
\end{figure}

\stitle{Finding 3. Charts and text play complementary roles across analytical tasks.}\quad To locate where each representation helps, we slice results by question type and cognitive load. Figure~\ref{fig:probe-slices} shows a clear pattern. Charts pull ahead on slices that demand visual pattern recognition, including temporal change, pattern recognition, and anomaly detection. Text pulls ahead on slices that demand precise reading, including numeric lookup, information location, and conditional verification. Free-style does not automatically inherit the stronger of the two on visual-pattern slices, so agents fail to switch to charts even where charts help most. Effective chart use therefore requires state-aware representation selection, and uniform access to both formats alone is not enough.

\stitle{Summary of \probe{}.}\quad \probe{} shows that current data agents understand charts far better than they use them, and that charts and text play complementary roles under different task conditions. This gap and this complementarity together motivate a mechanism that learns state-dependent representation choice as part of the analytical policy, which we develop in Section~\ref{sec:method}.
\vspace{-2pt}

\FloatBarrier
\section{Learning to Use Charts through Experience}
\label{sec:method}

\begin{figure}[!t]
	\centering
	\includegraphics[width=0.9\linewidth]{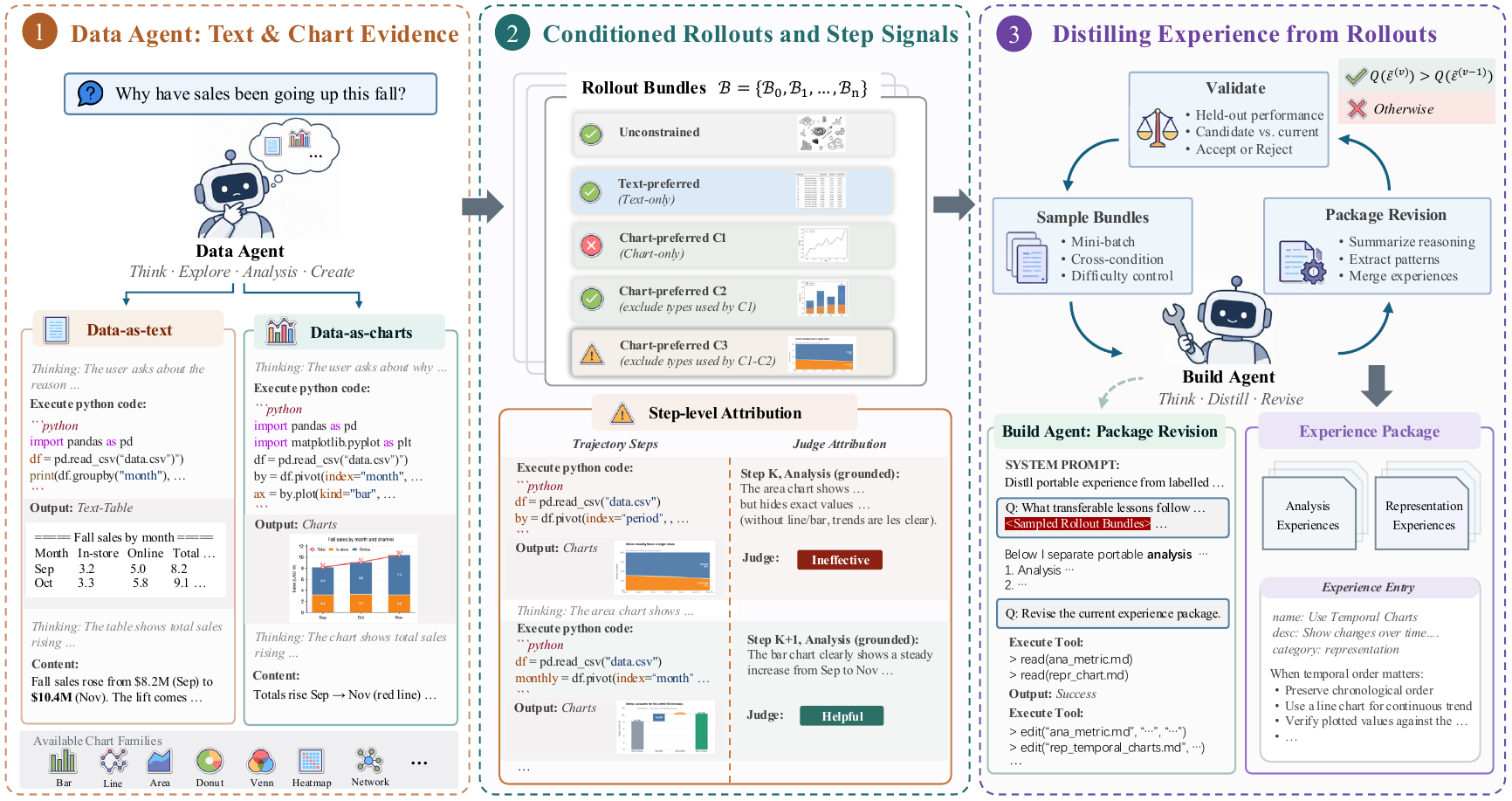}
		\caption{Overall pipeline of \method{}. Conditioned rollouts diversify representation preferences over each \probe{}. A trajectory judge produces outcome scores and step-level attributions. A builder agent distils analysis and representation experience under a validation gate.}
	\label{fig:method}
	\vspace{-4pt}
\end{figure}

\noindent Section~\ref{sec:probe} leaves a concrete target. Charts already carry sufficient evidence, yet agents do not invoke them where they help, so representation choice has to become a decision the agent makes from the current analysis state. Reaching that target requires three things. The agent needs a place to hold representation knowledge, a way to acquire that knowledge from its own behaviour, and a way to consult it while analysing. \method{} addresses these three needs in order. Section~\ref{sec:experience-memory} stores experience as editable documents that keep analytical and representational knowledge side by side. Section~\ref{sec:experience-distillation} distils those documents from conditioned rollouts under outcome and step-level supervision, with a validation gate on every update. Section~\ref{sec:online} lets the data agent consult the frozen documents inside its own analysis loop. Figure~\ref{fig:method} shows the pipeline. Base model parameters stay unchanged throughout.

\subsection{Storing Experience as Editable Documents}
\label{sec:experience-memory}

\noindent Representation choice depends on the analysis state, so the knowledge that governs it has to stay inspectable and revisable after training ends. We therefore keep experience outside the model as a document store that the agent reads during analysis and a builder revises during distillation. This choice also lets us audit what the agent learned and freeze it before any external evaluation.

\stitle{Two-layer storage.}\quad To make experience readable, editable, and callable on demand, we organise it as a two-layer package. The storage layer is a filesystem tree in which an entry document indexes independent experience documents. Each document carries metadata fields \texttt{name}, \texttt{description}, and \texttt{category} together with its content, which supports direct reading, targeted editing, and versioning. The tool layer exposes \texttt{Read}, \texttt{Create}, \texttt{Edit}, and \texttt{Bash} for managing the package. The builder agent uses all four, while the data agent accesses experience only through a read-only retrieval interface, implemented as \texttt{kb\_read} in downstream experiments. The data agent first fetches document metadata and then reads relevant content on demand, so context load stays low and grows with task complexity.

\stitle{Analysis and representation experience.}\quad To separate what to compute from how to represent it, we index two experience families under different categories. Analysis experience $\mathcal{E}_{\mathrm{ana}}$ holds reusable rules for selecting, organising, and verifying analytical operations. Representation experience $\mathcal{E}_{\mathrm{repr}}$ holds rules for when intermediate evidence should be represented as text, charts, or a mixture, and for how to consume the resulting evidence in later steps. The two families share the same storage and tool interface, and they are always updated and validated as one package. We write the $v$-th accepted version as $\mathcal{E}^{(v)}=\bigl(\mathcal{E}_{\mathrm{ana}}^{(v)},\mathcal{E}_{\mathrm{repr}}^{(v)}\bigr)$.

\subsection{Distilling Experience from Conditioned Rollouts}
\label{sec:experience-distillation}

\noindent A single rollout per task reveals only the habit the backbone already has, and Section~\ref{sec:probe-findings} shows that this habit is text-heavy for most backbones. Distillation from such data would reproduce the gap we want to close. We therefore force each task to be solved several times under different representation conditions, so the corpus contains chart-led trajectories even for backbones that would never produce them on their own, and outcome differences become attributable to representation choice.

\stitle{Conditioned rollout bundle.}\quad To realise this, we generate multiple trajectories per task while holding the base agent, tools, and interaction budget fixed. Let $p_i=(D_i,q_i,z_i,y_i^{\star})$ denote the $i$-th \probe{}, where the four elements are the data, question, task context, and reference answer. A trajectory generated under representation condition $\rho\in\{\mathrm{U},\mathrm{T},\mathrm{C}_1,\mathrm{C}_2,\mathrm{C}_3\}$ is denoted by $\tau_i^{\rho}$, and each task yields a bundle
\begin{equation}
\mathcal{B}_i=\bigl\{\tau_i^{\mathrm{U}},\tau_i^{\mathrm{T}},\tau_i^{\mathrm{C}_1},\tau_i^{\mathrm{C}_2},\tau_i^{\mathrm{C}_3}\bigr\},
\end{equation}
where $\tau_i^{\mathrm{U}}$ is an unconstrained rollout, $\tau_i^{\mathrm{T}}$ is a text-preferred rollout, and $\{\tau_i^{\mathrm{C}_k}\}_{k=1}^{3}$ are three chart-preferred rollouts. To keep the chart-preferred rollouts from collapsing onto a single visual habit, we enforce type exclusion across them. The second rollout cannot reuse any chart type used by the first, and the third cannot reuse any type used by the first two. This diversifies visual analyses without presuming that any chart type is intrinsically superior.

\stitle{Outcome and step signals.}\quad Bundle-level outcomes alone would tell the builder which trajectory performed best but not which decisions contributed to its performance, so we attach two labels to every trajectory. The outcome scoring function $f_{\mathrm{score}}(\tau_i^{\rho}\mid q_i,y_i^{\star})\!\in\![0,1]$ evaluates the final answer against the reference. The step-level attribution function $f_{\mathrm{attr}}(\tau_i^{\rho}\mid q_i,y_i^{\star})$ labels each key step with a tag in $\{\texttt{Helpful},\texttt{Ineffective},\texttt{Harmful}\}$ and separates the contribution of analytical operations from that of representation choices. Together the two functions tell the builder which behaviours to retain and which to avoid.

\stitle{Insight induction and package revision.}\quad To turn labelled trajectories into transferable rules, we split \probe{} into construction and validation sets so that all trajectories from one task remain in the same split. At each iteration, we sample $b$ tasks from the construction set and feed their bundles and labels to the builder. The builder first performs \textit{insight induction}, comparing trajectories and their attributions to identify transferable analytical and representational patterns. It then performs \textit{package revision}, reading the current package $\mathcal{E}^{(v-1)}$ and either merging new patterns into existing documents or creating fresh entries.

\stitle{Validation-gated update.}\quad An unconstrained builder can overwrite a working document with a pattern that held only for a single batch, so every proposal must demonstrate transferable value. Let $Q(\mathcal{E})$ denote validation performance with package $\mathcal{E}$, and let $\widetilde{\mathcal{E}}^{(v)}$ be the proposal at iteration $v$. The update rule is
\begin{equation}
\mathcal{E}^{(v)}=
\begin{cases}
\widetilde{\mathcal{E}}^{(v)}, & Q(\widetilde{\mathcal{E}}^{(v)})>Q(\mathcal{E}^{(v-1)}),\\[2pt]
\mathcal{E}^{(v-1)}, & \text{otherwise}.
\end{cases}
\label{eq:experience-gate}
\end{equation}
After multiple iterations, the accepted package $\mathcal{E}^{\star}$ is frozen for downstream use. The gate keeps validation performance monotonically nondecreasing and filters proposals that fit a single batch.

\subsection{Consulting Experience inside the Analysis Loop}
\label{sec:online}

\noindent A frozen package is useful only if the agent consults it when a representation decision arises. We therefore attach the package to the running analysis loop as a read-only resource, so no extra training, retrieval model, or change to the core ReAct execution loop is needed.

\stitle{Access to the frozen package.}\quad The data agent treats $\mathcal{E}^{\star}$ as external memory. It first receives document metadata, then reads the entries that match the question and the current analysis state. Reading costs a few tool calls and leaves the backbone untouched, so the same package serves every backbone we evaluate.

\stitle{Representation choice at each step.}\quad Analysis experience $\mathcal{E}_{\mathrm{ana}}^{\star}$ guides operation selection and verification, and representation experience $\mathcal{E}_{\mathrm{repr}}^{\star}$ guides whether an intermediate result should be represented as text, charts, or both. Charts therefore enter the loop when the analysis state calls for visual pattern recognition, instead of appearing only as a final artefact.

\section{Experiments}
\label{sec:experiments}

\noindent This section evaluates \method{} on three held-out benchmarks that cover analytical report generation, long-horizon exploration, and exact-answer analysis. We describe the benchmarks and evaluation protocol, report main results by capability, and analyse the mechanism through ablations.

\subsection{Benchmarks}
\label{sec:benchmarks}

\noindent To cover the main task forms of agentic data analysis, we adopt three benchmarks with complementary demands and use each benchmark's official protocol.

\stitle{Analytical report generation.}\quad \textbf{DAComp-DA} \citep{lei2025dacomp} contains open-ended tasks over potentially multi-table SQLite databases and requires reports with analysis, conclusions, and visualisations. It is evaluated by Completeness, Accuracy, Insightfulness, Readability, Analytical Depth, Visualisation, and Overall.

\stitle{Long-horizon exploration.}\quad \textbf{DDR-Bench} \citep{liu2026ddrbench} evaluates long-horizon exploration without predefined queries in a 10-K setting over SEC disclosures. It is evaluated by Message-Wise accuracy, Trajectory-Wise accuracy, and Overall.

\stitle{Exact-answer analysis.}\quad \textbf{DSBench} \citep{jing2024dsbench} derives exact-answer analytical tasks from ModelOff competitions with spreadsheet, text, and image inputs. It is evaluated by Task-level Accuracy, Competition-level Accuracy, and Overall.

\begin{table}[t]
	\caption{Analytical report generation on DAComp-DA. Higher is better on all metrics. \method{} shares every configuration with the matched baseline except access to $\mathcal{E}^{\star}$.}
	\label{tab:dacomp-matrix}
	\centering
	\small
	\setlength{\tabcolsep}{3pt}
	\begin{tabular}{@{}llccccccc@{}}
		\toprule
		Method & LLM Backbone & Comp. & Acc. & Insi. & Read. & Depth & Vis. & Overall \\
		\midrule
		\multirow{2}{*}{OpenHands} & GPT-5 & 60.98 & 40.30 & 49.39 & 35.51 & 69.80 & 21.40 & 46.99 \\
		& DeepSeek-V3.1 & 49.88 & 33.25 & 41.66 & 36.00 & 33.20 & 11.00 & 33.87 \\
		\midrule
		\multirow{2}{*}{DA-Agent} & GPT-5 & 64.23 & 43.81 & 56.89 & 43.59 & 76.80 & 27.44 & 50.84 \\
		& DeepSeek-V3.1 & 48.74 & 32.97 & 42.43 & 37.25 & 35.00 & 11.45 & 34.33 \\
		\midrule
		\multirow{4}{*}{Baseline}
		& Claude Sonnet 5 & 70.27 & 52.55 & 68.85 & 90.80 & 89.00 & 62.80 & 69.84 \\
		& GPT-5.6 Luna & 66.21 & 46.61 & 65.45 & 83.83 & 80.74 & 7.66 & 54.32 \\
		& GPT-5.6 Sol & 71.74 & 58.04 & 72.42 & 94.60 & 93.00 & 61.80 & 73.24 \\
		& DeepSeek V4.1 Flash & 83.91 & 67.80 & 81.77 & 98.00 & 94.44 & 75.72 & 82.23 \\
		\midrule
		\multirow{4}{*}{\method{}}
		& Claude Sonnet 5 & 73.31 & 54.12 & 70.20 & 91.17 & 89.74 & 68.91 & 72.37 \\
		& GPT-5.6 Luna & 68.10 & 51.18 & 65.24 & 82.83 & 85.00 & 27.61 & 59.78 \\
		& GPT-5.6 Sol & 72.38 & 60.32 & 74.08 & 95.94 & 93.54 & 63.03 & 75.81 \\
		& DeepSeek V4.1 Flash & 84.11 & 70.41 & 84.36 & 98.10 & 95.03 & 78.34 & 83.94 \\
		\bottomrule
	\end{tabular}
\end{table}

\begin{table}[t]
	\centering
	\begin{minipage}[t]{0.49\textwidth}
		\centering
		\captionof{table}{Long-horizon exploration on DDR-Bench (10-K). Msg. and Traj. denote Message-Wise and Trajectory-Wise accuracy.}
		\label{tab:ddrbench-matrix}
		\scriptsize
		\setlength{\tabcolsep}{2pt}
		\resizebox*{\linewidth}{140pt}{%
			\begin{tabular}{@{}llccc@{}}
				\toprule
				Method & LLM Backbone & Msg. & Traj. & Overall \\
				\midrule
				\multirow{4}{*}{ReAct}
				& Claude Sonnet 4.5 & 77.61 & 60.58 & 69.10 \\
				& GPT-5.1 & 37.12 & 44.25 & 40.69 \\
				& GPT-5.2 & 44.89 & 41.09 & 42.99 \\
				& DeepSeek V3.2 & 60.08 & 38.15 & 49.12 \\
				\midrule
				\multirow{4}{*}{Baseline}
				& Claude Sonnet 5 & 61.19 & 52.18 & 56.69 \\
				& GPT-5.6 Luna & 44.52 & 45.11 & 44.82 \\
				& GPT-5.6 Sol & 60.20 & 58.89 & 59.55 \\
				& DeepSeek V4.1 Flash & 64.19 & 59.01 & 61.60 \\
				\midrule
				\multirow{4}{*}{\method{}}
				& Claude Sonnet 5 & 64.01 & 60.31 & 62.16 \\
				& GPT-5.6 Luna & 47.28 & 49.62 & 48.45 \\
				& GPT-5.6 Sol &  63.25 & 59.25 &  61.25\\
				& DeepSeek V4.1 Flash & 65.13 & 60.43 & 62.77 \\
				\bottomrule
			\end{tabular}
		}
	\end{minipage}\hfill
	\begin{minipage}[t]{0.49\textwidth}
		\centering
		\captionof{table}{Exact-answer analysis on the DSBench Data Analysis subset. Task and Comp. denote Task-level and Competition-level accuracy.}
		\label{tab:dsbench-matrix}
		\scriptsize
		\setlength{\tabcolsep}{2pt}
		\resizebox*{\linewidth}{140pt}{%
			\begin{tabular}{@{}llccc@{}}
				\toprule
				Method & LLM Backbone & Task & Comp. & Overall \\
				\midrule
				\multirow{2}{*}{Code Interpreter}
				& GPT-3.5 & 11.16 & 8.23 & 9.70 \\
				& GPT-4o & 23.82 & 22.65 & 23.24 \\
				\midrule
				\multirow{2}{*}{AutoGen}
				& GPT-3.5 & 20.82 & 13.80 & 17.31 \\
				& GPT-4o & 34.12 & 26.72 & 30.42 \\
				\midrule
				\multirow{4}{*}{Baseline}
				& Claude Sonnet 5 & 88.47 & 88.42 & 88.45 \\
				& GPT-5.6 Luna & 75.11 & 73.84 & 74.47 \\
				& GPT-5.6 Sol & 89.90 & 89.40 & 89.70 \\
				& DeepSeek V4.1 Flash & 91.32 & 90.64 & 90.98 \\
				\midrule
				\multirow{4}{*}{\method{}}
				& Claude Sonnet 5 & 90.31 & 89.93 & 90.12 \\
				& GPT-5.6 Luna & 76.72 & 76.25 & 76.49 \\
				& GPT-5.6 Sol &  89.94 & 89.65  & 89.80 \\
				& DeepSeek V4.1 Flash & 91.43 & 90.93 & 91.18 \\
				\bottomrule
			\end{tabular}
		}
	\end{minipage}
\end{table}

\subsection{Experimental Setup}
\label{sec:main-exp-setup}

\stitle{Matched agent stack.}\quad The baseline agent and \method{} share the model version, decoding configuration, base system instructions, data access, interaction budget, ReAct loop, and analytical tools within each backbone. \method{} additionally receives the experience-access instructions, catalogue metadata, and read-only \texttt{kb\_read} tool required to consult the frozen package.

\stitle{Leakage prevention.}\quad All experience distillation uses \probe{} only, and the three benchmarks are held out from experience construction, so downstream numbers measure transfer. Reported prior results are compared only when the split and protocol match.

\stitle{Package freezing.}\quad We build a single package with GPT-5.6 Sol and freeze the accepted version before external evaluation. The same frozen package is then used by every evaluated backbone, so no downstream evaluation signal reaches the package.

\subsection{Main Results}
\label{sec:main-results}

\stitle{Analytical report generation.}\quad Table~\ref{tab:dacomp-matrix} shows that \method{} consistently improves report-style analysis across all four matched backbones, raising Overall by 1.71--5.46 points, with a mean gain of 3.07. The largest improvement appears in Visualisation, which increases by 7.48 points on average and by 19.95 points for GPT-5.6 Luna. Accuracy also rises for every backbone, by 2.76 points on average. These coupled gains indicate that the learned experience improves both analytical correctness and evidence presentation, rather than merely producing more polished reports.

\stitle{Long-horizon exploration.}\quad On DDR-Bench, \method{} improves Overall for every matched backbone by 1.17--5.47 points, yielding a mean gain of 2.99 points. Message-Wise and Trajectory-Wise accuracy increase consistently, by 2.39 and 3.60 points on average, respectively. The larger trajectory-level gains, particularly for Claude Sonnet 5 and GPT-5.6 Luna, show that the benefit extends beyond isolated responses and accumulates over the full analytical trajectory. Thus, the learned experience supports both local analytical decisions and coherent long-horizon exploration.

\stitle{Exact-answer analysis.}\quad Table~\ref{tab:dsbench-matrix} shows consistent gains on DSBench despite the strong baseline performance. \method{} improves Overall for all four backbones by 0.10--2.02 points, with a mean gain of 1.00 point. Task-level and Competition-level accuracy increase by 0.90 and 1.12 points on average, respectively. Gains are larger for Claude Sonnet 5 and GPT-5.6 Luna, while the smaller improvements on GPT-5.6 Sol and DeepSeek V4.1 Flash likely reflect their limited remaining headroom. These results demonstrate that richer analytical experience does not trade off exact-answer correctness.

\begin{table}[t]
\centering
\begin{minipage}[t]{0.52\textwidth}
\centering
\captionof{table}{Distillation procedure on DAComp-DA.}
\label{tab:ablation-construction}
\small
\setlength{\tabcolsep}{5pt}
\begin{tabular}{@{}lccc@{}}
\toprule
Variant & Acc. & Vis. & Overall \\
\midrule
baseline & 46.61 & 7.66 & 54.32 \\
one-step distillation & 49.31 & 18.75 & 56.94 \\
w/o conditioned rollouts & 47.43 & 8.43 & 57.11 \\
w/o validation gate & 46.93 & 12.43 & 56.73 \\
\method{} & 51.18 & 27.61 & 59.78 \\
\bottomrule
\end{tabular}
\end{minipage}\hfill
\begin{minipage}[t]{0.44\textwidth}
\centering
\captionof{table}{Experience types on DAComp-DA.}
\label{tab:ablation-experience-types}
\small
\setlength{\tabcolsep}{5pt}
\begin{tabular}{@{}lccc@{}}
\toprule
Variant & Acc. & Vis. & Overall \\
\midrule
baseline & 46.61 & 7.66 & 54.32 \\
analysis only & 50.31 & 10.43 & 58.74 \\
representation only & 49.31 & 20.75 & 57.31 \\
\method{} & 51.18 & 27.61 & 59.78 \\
\bottomrule
\end{tabular}
\end{minipage}
\end{table}

\subsection{Ablation Studies and Mechanism Analysis}
\label{sec:ablations}

\stitle{Ablation protocol.}\quad To locate which mechanism inside \method{} drives the gains, we run all ablations on DAComp-DA with the same backbone, ReAct loop, tool interface, interaction budget, and evaluator. We report Acc., Vis., and Overall, which measure analytical correctness, visual evidence quality, and aggregate performance. Each ablation changes only the component under study.

\FloatBarrier

\begin{wrapfigure}[15]{r}{0.42\textwidth}
\centering
\includegraphics[width=\linewidth]{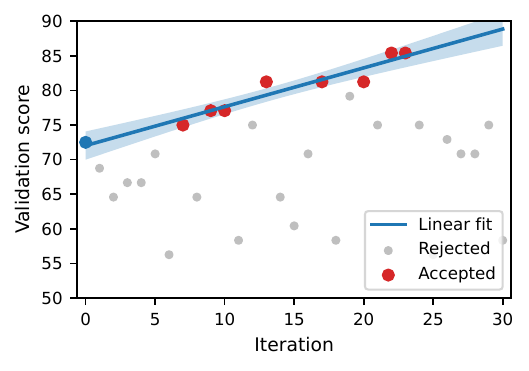}
\caption{Validation trajectory during experience distillation. The gate admits only proposals that improve held-out performance.}
\label{fig:experience-convergence}
\end{wrapfigure}

\stitle{Iterative distillation.}\quad One-step distillation improves Overall from 54.32 to 56.94 over the experience-free baseline, confirming the value of trajectory-derived experience. \method{} further reaches 59.78, exceeding one-step distillation by 1.87 points in Acc. and 8.86 points in Vis. Iterative refinement therefore provides additional value beyond a single-pass summary, particularly for constructing visual evidence.

\stitle{Conditioned rollouts and validation gate.}\quad Using only unconstrained rollouts reduces Overall from 59.78 to 57.11, while removing the validation gate lowers it to 56.73. Both variants remain above the baseline but fall short of full \method{}. Their larger drops in Vis. than in Acc. show that diverse representation-conditioned evidence and validation-based selection are particularly important for learning representation-aware analytical behaviour.

\stitle{Two experience families.}\quad Both experience families improve the baseline, but in different ways. Analysis experience raises Acc. by 3.70 points and achieves the stronger single-family Overall, whereas representation experience produces a larger 13.09-point gain in Vis. Combining them yields the best Acc., Vis., and Overall, showing that analytical operation selection and evidence representation provide complementary rather than redundant guidance.

\stitle{Optimisation dynamics.}\quad Figure~\ref{fig:experience-convergence} tracks the retained validation score across 30 candidate updates. The gate accepts eight proposals and raises the score monotonically from 72.50 to 85.42, while rejecting 22 others with scores as low as 56.25. Not every local edit transfers, and the gate stabilises optimisation by preventing harmful updates from entering the final package.

\begin{figure}[!t]
\centering
\includegraphics[width=0.75\linewidth,keepaspectratio]{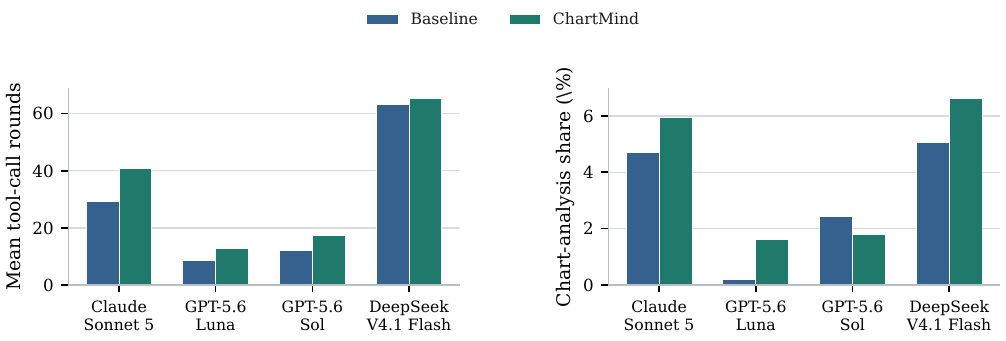}
\caption{DAComp-DA trajectory statistics for the matched baseline and \method{}. \textbf{Left:} mean tool-call rounds per completed task. \textbf{Right:} fraction of analysis-phase Bash calls (excluding the final two per task) whose observation includes a valid chart image.}
\label{fig:dacomp-trajectory-behaviour}
\end{figure}

\stitle{Trajectory behaviour on DAComp-DA.}\quad To connect the ablations to agent behaviour rather than only judge scores, we aggregate completed trajectories on DAComp-DA for the matched baseline and full \method{} runs. For each backbone we report the mean number of tool-call rounds per task and the share of \emph{analysis-phase} Bash calls---all but the final two invocations per task---that emit a valid chart image. Figure~\ref{fig:dacomp-trajectory-behaviour} shows a modest increase in interaction depth under \method{}, reflecting the additional rounds used to consult the experience library and conduct more thorough analytical exploration. On three of the four backbones, \method{} also increases analysis-phase plotting frequency, with GPT-5.6 Luna showing the clearest shift from rarely plotting under the baseline to using charts during analysis. Together with the Vis.\ gains across all four backbones in Table~\ref{tab:dacomp-matrix}, these results indicate that \method{} generally encourages more frequent in-loop chart use while consistently improving the quality of visual evidence.

\vspace{8pt}

\FloatBarrier
\section{Related Work}
\label{sec:related}

\stitle{Data agents.}\quad LLM-driven data agents follow a plan--execute--observe--revise loop over structured data. Representative systems couple reasoning with executable environments and cover heterogeneous analytical tasks. DS-Agent \citep{guo2024dsagent} retrieves historical cases inside an executable loop, Data Interpreter \citep{hong2025datainterpreter} organises tasks as dynamic directed acyclic graphs, AIDE \citep{jiang2025aide} models solution development as tree search over code, DS-STAR \citep{nam2025dsstar} adds sufficiency verification, AutoKaggle \citep{li2025autokaggle} adopts multi-agent division of labour, DeepAnalyze \citep{zhang2025deepanalyze} trains an end-to-end data science model, and InsightBench \citep{sahu2025insightbench} extends evaluation to open-ended business insight generation. These systems keep the analytical loop text-centric and treat charts primarily as report artefacts. \method{} instead makes charts first-class intermediate actions in the analytical loop.

\stitle{Chart generation and understanding.}\quad One line of work optimises the production or reading of a single chart. LIDA \citep{dibia2023lida}, MatPlotAgent \citep{yang2024matplotagent}, AMACE \citep{namgoong2025amace}, MultiVis-Agent \citep{lu2026multivisagent}, and ChartOptimiser \citep{wang2025chartoptimiser} improve chart generation through rendered feedback, multi-agent collaboration, and intent-aware design, while ChartMimic \citep{yang2025chartmimic} benchmarks chart-to-code generation. UniChart \citep{masry2023unichart} and ChartLlama \citep{han2023chartllama} advance chart reading through specialised training, while ChartBench \citep{xu2023chartbench} benchmarks complex chart reasoning. DePlot \citep{liu2023deplot}, MatCha \citep{liu2023matcha}, and TinyChart \citep{zhang2024tinychart} strengthen numerical analysis through derendering and programmatic reasoning. These methods assume that the chart or the plotting request is already fixed. \method{} decides when a chart should exist at all inside an analytical trajectory.

\stitle{Representation choice for reasoning.}\quad Recent work studies how textual, tabular, or visual representations are used in reasoning. ChartAgent \citep{kaur2026chartagent} and Charts-of-Thought \citep{das2025chartsofthought} reason over existing charts, FRES \citep{zhou2025fres} and FormatMatters \citep{ho2026formatmatters} compare different input representations, TableDART \citep{xing2026tabledart} routes table--query pairs across textual, visual, and fused paths, and EvidFuse \citep{lin2026evidfuse} builds visual evidence on demand during report writing. \method{} learns state-dependent representation choice from labelled analytical trajectories, so text and chart evidence are selected as the analysis unfolds.

\section{Conclusion}
\label{sec:conclusion}

\noindent We study whether charts can serve as internal epistemic actions inside a data agent's analytical loop. \probe{} shows that current agents extract sufficient evidence from self-generated charts, yet plotting access alone does not translate into consistent chart use. Charts and text play complementary roles across task slices, with charts stronger on temporal, pattern, and anomaly reasoning and text stronger on precise reading and conditional verification. Motivated by these findings, \method{} distils analysis experience $\mathcal{E}_{\mathrm{ana}}$ and representation experience $\mathcal{E}_{\mathrm{repr}}$ from conditioned trajectory rollouts using outcome scoring $f_{\mathrm{score}}$ and step-level attribution $f_{\mathrm{attr}}$, and retains only experience updates that pass a validation gate. On three held-out benchmarks with four LLM backbones, \method{} induces in-loop chart use, improves analytical accuracy, and lifts visual evidence quality without touching model parameters. These results position representation choice as part of the analytical policy rather than as a cosmetic output stage.



\bibliographystyle{iclr2027_conference}
\bibliography{iclr2027_conference}

\clearpage
\appendix

\section{Token Cost}
\label{app:additional-results}
\label{app:construction-tokens}

\stitle{Conditioned rollout cost.}\quad
Table~\ref{tab:rollout-token-cost} reports API-recorded prompt and completion
tokens for the rollout bundle and its post-rollout judge. With
\texttt{gpt-5.6-sol}, 174 training tasks produce 870 trajectories: one
unconstrained, one text-preferred, and three chart-preferred rollouts per task.

\begin{table}[H]
\caption{Token cost of conditioned rollout collection on the train split
(174 tasks for each of the unconstrained and text-preferred policies; chart-preferred aggregates three runs per task).}
\label{tab:rollout-token-cost}
\centering
\small
\setlength{\tabcolsep}{3.5pt}
\begin{tabular}{@{}l r r r r r r@{}}
\toprule
Rollout policy & Traj. &
\multicolumn{2}{c}{Rollout agent} &
\multicolumn{2}{c}{Trajectory judge} &
Total \\
\cmidrule(lr){3-4}\cmidrule(lr){5-6}
& & Input & Output & Input & Output & \\
\midrule
Unconstrained       & 174 & \num{917194}  & \num{151749}  & \num{738012}  & \num{111615}  & \num{1918570} \\
Text-preferred      & 174 & \num{938777}  & \num{167501}  & \num{747263}  & \num{112236}  & \num{1965777} \\
Chart-preferred     & 522 & \num{3489026} & \num{623247}  & \num{2920261} & \num{357221}  & \num{7389755} \\
\midrule
All train rollouts  & 870 & \num{5344997} & \num{942497}  & \num{4405536} & \num{581072}  & \num{11274102} \\
\bottomrule
\end{tabular}
\end{table}

\stitle{Experience distillation cost.}\quad
Table~\ref{tab:distill-token-summary} aggregates the rollout corpus, builder,
validation-probe agent, and probe judge over 30 distillation iterations with 12
validation probes per iteration. All counts are read from \texttt{llm\_log}.

\begin{table}[H]
\caption{Total token cost of experience construction. Distillation uses 30
iterations with 12 validation probes per iteration.}
\label{tab:distill-token-summary}
\centering
\small
\setlength{\tabcolsep}{4pt}
\begin{tabular}{@{}l r r r@{}}
\toprule
Stage & Input & Output & Total \\
\midrule
Train rollout corpus (Table~\ref{tab:rollout-token-cost}) &
\num{9750533} & \num{1523569} & \num{11274102} \\
Builder (30 iterations) &
\num{9237636} & \num{174256} & \num{9411892} \\
Gate probe agent &
\num{4235416} & \num{387437} & \num{4622853} \\
Gate trajectory judge &
\num{2160964} & \num{226357} & \num{2387321} \\
\midrule
Full construction pipeline &
\num{25384549} & \num{2311619} & \num{27696168} \\
\bottomrule
\end{tabular}
\end{table}

\label{app:downstream-tokens}

\stitle{Downstream evaluation cost.}\quad
Table~\ref{tab:downstream-token-cost} reports mean prompt, completion, and total
tokens per completed downstream task for GPT-5.6 Luna, comparing the matched
data agent without and with the frozen experience library. Counts come from
each task's \texttt{trajectory.json}; benchmark judges and offline evaluators
are excluded.

\begin{table}[H]
\caption{Mean agent token cost per completed task for GPT-5.6 Luna, without
and with the frozen experience library.}
\label{tab:downstream-token-cost}
\centering
\small
\setlength{\tabcolsep}{4pt}
\begin{tabular}{@{}l rrr rrr@{}}
\toprule
Benchmark &
\multicolumn{3}{c}{Baseline (no KB)} &
\multicolumn{3}{c}{\method{} (with KB)} \\
\cmidrule(lr){2-4}\cmidrule(lr){5-7}
& Input & Output & Total & Input & Output & Total \\
\midrule
DAComp-DA & \num{73339} & \num{9312} & \num{82651}
& \num{122451} & \num{9733} & \num{132184} \\
DDR-Bench & \num{89797} & \num{10923} & \num{100720}
& \num{111517} & \num{8483} & \num{120000} \\
DSBench & \num{40262} & \num{1614} & \num{41876}
& \num{68476} & \num{1658} & \num{70134} \\
\bottomrule
\end{tabular}
\end{table}

\clearpage

\section{Implementation Details and Core Prompts}
\label{app:implementation}

We reproduce the core prompt text used by the implementation in Markdown
format. Every displayed instruction is copied verbatim from the code;
angle-bracket fields denote run-time values, and square-bracket notes mark
omitted repetitive inputs, catalogues, or output schemas.

\subsection{Baseline Data Agents}
\label{app:data-agents}

\stitle{Data agent without an experience library.}\quad
The baseline uses a ReAct loop with \texttt{execute\_bash} and three file tools.
The benchmark instruction is passed unchanged as the user message. The core
system prompt is:

\begin{promptblock}{System prompt: data agent without experience}
You are a professional data analyst working on an open-ended business question over a SQLite database.
The database path is available as $SQLITE_PATH in Bash. You may:
- execute_bash for Python/sqlite3 analysis
- read_file / write_file / edit_file to maintain the Markdown report

Rules:
1. Do not clean the data; focus on analysis and actionable recommendations.
2. Do not train machine learning models.
3. Explore this task's SQLite with execute_bash (schema, samples, and queries/aggregates tied to the question).
4. Write the full report to `<instance_id>.md` using write_file and refine with edit_file.
5. You may create charts via Bash and reference them with relative Markdown image paths.
6. Finish with a brief natural-language summary without further tool calls.

Maximum <max_turns> tool turns. All artifacts must be in English.
\end{promptblock}

\clearpage

\stitle{Data agent with an experience library.}\quad
\method{} keeps the same model, task input, ReAct loop, budget, and report
procedure, and adds only the read-only \texttt{kb\_read} tool and catalogue. The
following excerpt shows the exact additions to the shared system prompt:

\begin{promptblock}{Core system-prompt additions: data agent with experience}
- kb_read to load one portable analysis or visualization experience from the catalog below (after you have explored this task's data)

3. You must explore this task's SQLite with execute_bash first (schema, samples, and queries/aggregates tied to the question). The experience library does not replace data exploration.

## KB exploration is not data exploration

Reading or browsing the experience library (`kb_read`, catalog below) is **not** data exploration. It does **not** substitute for inspecting and querying this task's actual dataset(s).

Before you answer, finish, or write the final report, you must **fully explore the data** for this question: available files/tables, schema, samples, row counts, filters/cohorts, aggregates, and sanity checks--via `execute_bash` (and file tools when maintaining reports). Ground every conclusion in what you computed on this task's data.

## Experience library (on-demand)

Workflow (required):
1. Data exploration first (see above)--never skip because an experience entry sounds relevant.
2. Only after you understand this task's data, optionally call `kb_read` if a catalog entry fits your *next* step (method or visualization)--then apply it with fresh queries on this data.
3. Do not answer from catalog text, memory, or generic templates.

Catalog entries are portable heuristics, not ground truth for this task. Use `filename` (preferred) or `name`; optional `category`. Metadata below is not full text--call `kb_read` to load an entry.

[Library metadata and repeated catalogue rows omitted.]
\end{promptblock}

\clearpage

\subsection{Rollout Bundle}
\label{app:rollout-agent}

\stitle{Conditioned rollout.}\quad
Each construction task has one unconstrained, one text-preferred, and three
chart-preferred rollouts. Inputs, backbone, and tools are fixed; only the
representation instruction changes. For chart run $k$, the final diversity
block lists $\mathcal{F}_{i,k}=\bigcup_{j<k}T(\tau_i^{C_j})$. The complete
condition-specific system instructions are:

\begin{promptblock}{System prompts: conditioned rollouts}
### Shared base
You are a data analyst. Analyze only the provided dataset and answer the analysis question. Give the final answer directly, with the key finding and concrete supporting evidence. Do not mention tools or the analysis process. Never access ground-truth or reference-answer files.

### Unconstrained condition
You may call execute_bash for analysis. Each call is independent. Stdout, stderr, and generated chart images are returned. Save charts as PNG, JPG, JPEG, or WebP in the current directory.

### Text-preferred condition
You may call execute_bash for analysis. Each call is independent. Stdout, stderr, and generated chart images are returned. Save charts as PNG, JPG, JPEG, or WebP in the current directory. You MUST prioritize TEXT as the primary analysis representation: tables, printed statistics, and stdout summaries should carry most of the reasoning. Charts are optional supporting evidence only.

### Chart-preferred condition
You may call execute_bash for analysis. Each call is independent. Stdout, stderr, and generated chart images are returned. Save charts as PNG, JPG, JPEG, or WebP in the current directory. You MUST prioritize CHARTS as the primary analysis representation: save clear PNG/JPG/WebP figures and rely on visual patterns before stating conclusions.

### Diversity suffix for chart runs 2 and 3
CHART DIVERSITY (mandatory): Prior chart trajectories on this task already used these chart families: <forbidden_chart_types>. You MUST NOT use any of those chart types in this run. Choose a different visualization family from your tool calls.
\end{promptblock}

The shared user prompt supplies the read-only dataset path, description, schema,
first three rows, and analysis question, then ends with the exact instruction
\emph{``Answer the analysis question directly.''}

\clearpage

\stitle{Judge and step signal.}\quad
\label{app:trajectory-judge}
The judge runs only after answer generation, so reference information cannot
affect the rollout. It sees the final answer and a compact trajectory summary,
scores semantic correctness from 0 to 4, and attributes each step. The
normalised outcome is $f_{\mathrm{score}}=s/4$.

\begin{promptblock}{User prompt: trajectory judge}
You are now acting as a strict evaluator (not as the analyst).

Rollout policy tag (prompt intent, not ground truth): <modality_id>

Question:
<question>

Dataset description:
<dataset_description>

Reference insight (ground truth):
<reference_insight>

Reference analysis code (ground truth):
```python
<reference_analysis_code>
```

Agent final answer:
<final_answer>

Trajectory summary (tool use, charts, bash previews):
<trajectory_summary_json>

Score core_result with an integer from 0 to 4:
- 4 (fully correct): directly answers the question and completely captures the reference insight's central entities, direction, and relationship, with no error that changes the conclusion.
[Definitions for scores 3, 2, and 1 omitted.]
- 0 (incorrect or absent): does not answer the question, gives a wrong or contradictory central conclusion, or provides only an action recommendation.

Judge semantic equivalence rather than wording overlap.

Return exactly one JSON object with ONLY these top-level keys (no markdown fences):
[JSON field definitions and writing-order details omitted.]

step_verdict meanings:
- Helpful: clearly supports the correct conclusion or necessary intermediate results
- Ineffective: redundant, off-track, or unused exploration, including steps that are harmless but contribute little
- Harmful: wrong entity/direction/aggregation or misleading representation

Do NOT include schema_version, probtask_id, rollout_policy, chart_run_index, final_score, or meta.
\end{promptblock}

\clearpage

\subsection{Experience Build}
\label{app:builder-agent}

\stitle{Builder agent.}\quad
The builder receives mini-batches of compact judge reports in two turns. The
first turn induces lessons without tools; the second retrieves overlapping
entries and applies the plan. Its tools are confined to the experience package.

\begin{promptblock}{System prompt: builder agent}
You distill a portable experience library for data-analysis agents from compact judge summaries.

Maintain two parallel tracks at all times:
- **analysis** -- filters, aggregations, validation, metric contracts, comparisons (same logic for text or chart).
- **representation** -- modality choice (text vs chart), when to plot, chart type, chart-then-verify, labels, and harmful visuals.

Rules:
- Corpus blocks are judge summaries (Q/In/Out, Why/Core, per-step A/R verdicts), not raw trajectories.
- Distill cross-task portable lessons; never memorize probtask ids, entities, or numeric answers from Out.
- Study helpful vs ineffective/harmful step attributions and score_reason; compare **text vs chart** runs on the same task for representation signal.
- Prefer high-quality atomic entries; merge overlapping ideas instead of duplicating.
- Follow each user message: planning turns are text-only; apply turns use repository tools.
\end{promptblock}

\clearpage

\begin{promptblock}{First user prompt: induction and planning}
# Task: distill portable experience (planning only)

You maintain a portable experience library for data-analysis agents. **This message is planning only** -- do **not** read or modify the experience repository (no tool calls). Put your entire answer in the assistant message.

## Judge summary blocks (below)

Each mini-batch is several rollout blocks separated by `---`. No raw trajectories and no ground-truth answer line -- infer lessons only from judge text.

**Policies (match corpus tags):** `base` (text + chart), `text` (text-first), `chart` (chart-oriented; often multiple runs `r0`, `r1`, ...). When the same probtask appears under different policies or runs, compare step verdicts and **A vs R** attribution across modalities -- that contrast is the main signal for **representation** lessons.

**Category routing:** mistakes about data, filters, metrics, denominators, or populations -> **analysis**, even on a chart run. Modality choice, chart type, labels, axes, chart-then-verify, decorative or misleading visuals -> **representation**.

## How to distill (what to write)

Extract cross-task patterns from `Why`, `Core`, and step verdicts -- not from memorizing answers or outputs.

Distill **analysis** and **representation** on separate tracks; every future file has exactly one `category`.

- **analysis** -- filters, aggregations, validation, metric contracts, comparisons, eligibility/denominators.
- **representation** -- how to present results: modality choice, when to plot, chart type, chart-then-verify, harmful visuals the judge flagged.

- **Wrong / low-score rollouts:** phrase as preventive rules ("when S, verify Y / avoid Z").
- **Correct / high-score rollouts:** phrase as reusable procedure ("when S, do Y"), not "for this dataset ...".
- **Do not store:** pasted `Out` text, numeric/chart conclusions from the case, dataset-specific facts.

## Output format

Reply with two sections:
### Analysis (candidate lessons)
### Representation (candidate lessons)

[Catalogue counts, gate probes, output details, and judge-summary corpus omitted.]
\end{promptblock}

\clearpage

\begin{promptblock}{Second user prompt: package revision}
# Task: update the experience repository

Use your tools to apply the plan from the previous assistant message. Changes must pass bundle validation.

## Experience file format

- One atomic experience per file. Filename **must** start with the same prefix as `category`: `references/analysis_<topic>.md` or `references/representation_<topic>.md` (`<topic>` = lowercase snake_case words only -- no hyphens or camelCase). The prefix must match frontmatter `category`. Rename with bash when creating, splitting, or fixing legacy names.
- YAML frontmatter plus body (or `content:` in frontmatter). Required keys: `name`, `description`, `category`.
- `category` is `analysis` or `representation` only; never mix both topics in one file.
- Body: non-empty markdown -- portable guidance, not task-specific answers. `description` must state what the rule is and when it applies in one line.

Tools: prefer `edit_experience` for small updates, `write_experience` for new or heavy rewrites; bash only for rename/delete under this skill directory.

## Required workflow

1. **Retrieve first:** scan the catalog for entries in the same `category` as each planned lesson; `read_experience` every file that might overlap -- not only the closest match. If `references/` is non-empty, finish this read pass before any write/edit/bash.
2. **Choose operations** (default refine when an existing entry already covers the idea; otherwise add, split, or delete):
   - **Refine** -- same idea exists; improve triggers, scope, or wording.
   - **Add** -- only after retrieval shows a clearly missing distinct idea.
   - **Split** -- one file bundles multiple ideas or categories.
   - **Delete** -- duplicate, misleading, obsolete, or too task-specific.
   - **No change** -- for a planned lesson only, when the library already covers it well enough.
   Apply every planned lesson that is not adequately covered; there is **no limit** on how many files or edits you may make in one batch.
3. Implement the plan, then briefly summarize: files read, operations taken, and why.

[Run-progress guidance, prior plan, and catalogue payload omitted.]
\end{promptblock}

\clearpage

\stitle{Validation gate.}\quad
Each candidate package is evaluated on the same held-out probes by an
unconstrained analysis agent with read-only experience access. The judge prompt
is identical to the trajectory-judge prompt above; the mean normalised score is
used in Equation~\ref{eq:experience-gate}. The analysis prompts are:

\begin{promptblock}{System prompt: validation-probe agent}
You are a data analyst. Analyze only the provided dataset and answer the analysis question. Give the final answer directly, with the key finding and concrete supporting evidence. Do not mention tools or the analysis process. Never access ground-truth or reference-answer files.

You may call execute_bash for analysis. Each call is independent. Stdout, stderr, and generated chart images are returned. Save charts as PNG, JPG, JPEG, or WebP in the current directory.

## Experience catalog

- **<experience_id>** (<category>): <name> -- <description>
<one line per catalogue entry>

Use the read_experience tool to load full text for an id before applying it.
\end{promptblock}

\begin{promptblock}{User prompt: validation-probe agent}
Dataset path:
$DATASET_PATH (read-only CSV, available inside each Bash call)

Dataset description:
<dataset_description>

Dataset schema:
<schema>

First 3 rows:
<sample>

Analysis question:
<question>

Answer the analysis question directly.
\end{promptblock}

\clearpage

\section{Learned Experience Cases}
\label{app:learned-experiences}

\noindent The following accepted entries illustrate the two experience families
consulted by downstream agents. We show two portable analysis rules and two
representation rules exactly in their stored Markdown form.

\begin{caseblock}{Analysis experience case 1}
name: Define the metric contract before calculation\\
description: When a request uses an ambiguous metric term, define its entity,
unit, population, time basis, and aggregation before computing it.\\
category: analysis\\[3pt]
Before calculating an ambiguous measure, state the metric contract: entity and
observational unit; eligible population and exclusions; numerator and
denominator; time anchor and comparison window; grouping levels, aggregation,
and weighting. Inspect actual field values before mapping semantic terms to
labels, and do not replace the requested outcome with a convenient proxy.
Before finalizing, reconcile every computed field and reported conclusion
against this contract.
\end{caseblock}

\begin{caseblock}{Analysis experience case 2}
name: Decompose the drivers of an elevated average\\
description: When explaining why a group average is high, separate category
frequency, category intensity, and contribution to the total.\\
category: analysis\\[3pt]
For each component category, report its record count, mean or median value, and
share of the group's total. Compare the focal group's component mix with peers,
then compare like-for-like component values across groups. When useful,
recompute the focal average after excluding or standardizing the suspected
high-value components. This separates composition effects from within-category
intensity and prevents confusing the largest category by count with the main
per-record driver.
\end{caseblock}

\clearpage

\begin{caseblock}{Representation experience case 1}
name: Prefer text or tables for compact results\\
description: When the result is a simple scalar, ranking, or small grouped
summary, use concise text or a labeled table unless a chart adds material
value.\\
category: representation\\[3pt]
Match presentation complexity to the analytical structure. A direct sentence
or compact table is often clearer for a few exact values, a short ranking, or a
binary comparison. Add a chart only when it improves perception of pattern,
spread, trend, or scale; plotting is not a quality requirement by itself.
\end{caseblock}

\begin{caseblock}{Representation experience case 2}
name: Verify every chart against computed results\\
description: After generating a chart, reconcile its data, ordering, units,
labels, and claims with the supporting table before presenting it.\\
category: representation\\[3pt]
Perform a chart-then-verify pass: compare plotted values, population, filters,
derived metric, and category order with the source table; check axis units,
scales, baselines, legends, and time order; confirm titles, reference lines, and
annotations; and ensure that the narrative states the same population, metric,
direction, and comparison. If validation changes the data or ordering,
recompute the source table and regenerate the chart rather than interpreting a
stale figure.
\end{caseblock}

\end{document}

%% file: math_commands.tex
\usepackage{amsmath,amsfonts,bm}

\def\eqref#1{equation~\ref{#1}}

\def\1{\bm{1}}

\DeclareMathAlphabet{\mathsfit}{\encodingdefault}{\sfdefault}{m}{sl}
\SetMathAlphabet{\mathsfit}{bold}{\encodingdefault}{\sfdefault}{bx}{n}

